\documentclass{article}

\usepackage[preprint]{neurips_2024}
\usepackage{fancyhdr}
\usepackage{floatrow}  
\usepackage{graphicx}  
\usepackage{booktabs}  
\usepackage{blindtext} 
\newfloatcommand{capbtabbox}{table}[][\FBwidth]

\usepackage[utf8]{inputenc} 
\usepackage[T1]{fontenc}    
\usepackage{hyperref}       
\usepackage{url}            
\usepackage{booktabs}       
\usepackage{amsfonts}       
\usepackage{nicefrac}       
\usepackage{microtype}      
\usepackage{algorithm}
\usepackage{algpseudocode}
\usepackage{amsmath}
\usepackage{graphicx}
\usepackage[table]{xcolor}
\usepackage{xurl}

\title{Target Strangeness: A Novel Conformal Prediction Difficulty Estimator
}

\author{%
  Alexis Bose, Jonathan Ethier,  Paul Guinand \\
  Communications Research Centre Canada
}

\begin{document}

\maketitle

\begin{abstract}
\textbf{This paper introduces Target Strangeness, a novel difficulty estimator for conformal prediction (CP) that offers an alternative approach for normalizing prediction intervals (PIs). By assessing how atypical a prediction is within the context of its nearest neighbours' target distribution, Target Strangeness can surpass the current state-of-the-art performance. This novel difficulty estimator is evaluated against others in the context of several conformal regression experiments.}
\end{abstract}

\section{Background and related work}
Conformal Prediction (CP) provides a robust framework for constructing prediction intervals with statistical guarantees, regardless of the underlying data distribution or the predictive model used. By relying on the assumption of data exchangeability, CP generates intervals that contain the true value with a predetermined confidence level, offering a powerful tool for uncertainty quantification in machine learning applications \citet{angelopoulos2021gentle}.

A critical aspect of Conformal Prediction (CP) is the use of difficulty estimators to normalize nonconformity measures, which can enhance the efficiency of prediction intervals. While several estimators have been proposed \cite{kato2023review}, there is potential for new approaches that leverage local data structures. In this work, we introduce Target Strangeness, a difficulty estimator that assesses the unlikeliness of a prediction given the distribution of the nearest neighbours.

Our main contribution is to present Target Strangeness as a viable difficulty estimator for CP, providing empirical evidence of its performance through regression experiments.

Conformal Prediction (CP) offers a principled approach to quantifying the uncertainty of model predictions by constructing prediction intervals (PIs). By generating multiple predictions across various coverage (significance) levels, CP enables us to understand the range within which the true values are likely to lie. To refine these prediction intervals and reduce their widths, we introduce a normalization factor \(\sigma_i\) that estimates the difficulty or uncertainty associated with each prediction:

\begin{equation}
\alpha_i = \frac{|y_i - \hat{y}_i|}{\sigma_i}
\end{equation}
where:

\begin{itemize}
    \item $\alpha_i$ is the normalized nonconformity measure for instance $i$,
    \item $y_i$ is the true value,
    \item $\hat{y}_i$ is the predicted value,
    \item $\sigma_i$ is the difficulty estimate, which may depend on the input features, the prediction, and the desired coverage (significance) level.
\end{itemize}

By normalizing the nonconformity measures, we account for varying levels of uncertainty across different instances, allowing for more tailored prediction intervals. 

\section{Target Strangeness difficulty estimator}
\label{sec:Target Strangeness Difficulty Estimator}

Nearest neighbour-based normalization was first introduced by \citet{papadopoulos2011regression}.
We extend the concept of estimating difficulty using the standard deviation of K-nearest neighbours' targets to Target Strangeness, which estimates difficulty based on the likelihood that a prediction, $\hat{y}$ belongs to the nearest neighbours' target distribution $\theta$.
\begin{equation}
\ell(\theta ; \hat{y}) = \ln \mathcal{L}_n(\theta ; \hat{y})
\end{equation}
The essence of the Target Strangeness difficulty estimator is that a prediction which does not belong to the target distribution of the nearest neighbours will be more uncertain and have a larger PI.
Kernel density estimation (KDE) was used for its flexibility in not assuming a predefined distribution and for its various efficient implementations. The Target Strangeness difficulty estimator is defined with a KDE as follows:

\begin{equation}
L(\hat{y}) = \frac{1}{k \cdot h} \sum_{i=1}^k K\left(\frac{\hat{y} - y_i}{h}\right)
\end{equation}
\text{where:}
\begin{itemize}
    \item \( L(\hat{y}) \) is the likelihood of \( \hat{y} \) given the distribution of \( y \) values from the k-nearest neighbours.
    \item \( \hat{y} \) is the value for which the likelihood is being estimated.
    \item \( y_i \) are the \( y \) values of the \( k \)-nearest neighbours.
    \item \( K \) is the Gaussian kernel function.
    \item \( h \) is the bandwidth (0.75), a positive number that controls the smoothness of the density estimate.
    \item \( k \) is the number of nearest neighbours considered in the estimate.
\end{itemize}
Target Strangeness difficulty estimation amounts to fitting the KDE, evaluating the probability density, and finally computing the strangeness measure.
Kernel Density Estimation:

\begin{equation}
\text{KDE Fit} : f(y) = \frac{1}{kh} \sum_{i=1}^{k} K\left(\frac{(\hat{y} - {y_i})^2}{h}\right)
\end{equation}

Evaluating Density:

\begin{equation}
\text{Log Density at } y: \log f(\hat{y}) = \log \left( \frac{1}{kh} \sum_{i=1}^{k} \exp\left(-\frac{(\hat{y} - {y_i})^2}{2h^2}\right) \right)
\end{equation}

Strangeness Measure:
\begin{equation}
\text{Strangeness} : \sigma = 1 - f(\hat{y})
\end{equation}

Finally, Algorithm 1 provides the implementation.
\begin{algorithm}[htbp]
    \caption{Calculate Target Strangeness}
    \begin{algorithmic}[1]
        \Procedure{Calculate Target Strangeness}{$\text{neighbour\_indexes}, \text{y\_hats}$}
            \State \textbf{Input:} groups of nearest neighbours, $\text{nn\_idx}$, and array of $y_{\text{hats}}$
            \State $\text{sigmas} \gets []$
            \For{$idx, nn\_idx$ \textbf{in} \text{enumerate}($\text{neighbour\_indexes}$)}
                \State \# Make a distribution of target-values in each neighbourhood
                \State $\text{kde} \gets \text{KernelDensity.fit}(\text{self.y[nn\_idx]})$
                \State \# Calculate $y_{\text{hat}}$ target strangeness
                \State $\text{log\_dens} \gets \text{kde.score\_samples}(\text{y\_hat[idx]})$
                \State $\text{prob\_dens} \gets \text{exp}(\text{log\_dens})$
                \State \# Bigger is stranger
                \State $\text{sigmas.append}(1 - \text{prob\_dens})$
            \EndFor
        \EndProcedure
    \end{algorithmic}
\end{algorithm}
Different KDE python libraries were examined and described in these results \citet{kdepy2023comparison}. The scikit-learn \citet{scikit-learn} KernelDensity implementation was selected, because other libraries did not offer a noticeable performance advantage when exploring up to 100 nearest neighbours. 

\section{Experimental Setup}

We used the inductive, or split, approach to creating conformal regressors, as outlined by \citet{papadopoulos2002inductive}, which circumvents the need for computationally intensive model retraining. This method involves splitting the training data into a proper training set and a calibration set. The model is trained on the former, while the latter is utilized for nonconformity score computation. This is a desirable approach when dealing with large datasets and complex models.

Our experiments utilize this split approach using the Conformal Classifiers Regressors and Predictive Systems (CREPES) framework \citep{crepes}. This framework incorporates Conformal predictive systems (CPS) which enhance conformal regressors by generating cumulative probability distributions, over potential target values \citet{vovk2020computationally}. This integration facilitates the integration and evaluation of novel difficulty estimators. 

The CP training set is first used to train the difficulty estimator. This estimator is then applied to the CP calibration set to obtain calibrated difficulty estimates. Next, we build a conformal regressor (CR) model using the calibration residuals along with these calibrated difficulty estimates. The CR model transforms point predictions into prediction intervals at a specified confidence level. To generate prediction intervals for the test set, we create difficulty estimates based on the predicted targets. These test set difficulty estimates, together with the confidence level, are then used by the CR to produce the prediction intervals.

Two regression problems are considered:

1. \textbf{Baseline Comparison}: A housing price prediction task serves as a benchmark the performance of Target Strangeness against existing difficulty estimators.

2. \textbf{Machine Learning-Based Path Loss (MLPL) Regression}: We assess the estimator's effectiveness in a wireless communication context by predicting path loss using external features.

The normalized conformal regressors employed in the CPSs were: \textit{norm\_std}, which normalizes the standard deviation of the targets of the K-nearest neighbours, and \textit{norm\_targ\_strg}, the Target Strangeness described in section \ref{sec:Target Strangeness Difficulty Estimator}. A different approach to generating PIs of varying sizes involves partitioning the object space into distinct, non-overlapping Mondrian categories \citep{crepes}, where predictions are grouped into bins for categorization. The Mondrian variants employed in the CPSs were: \textit{norm\_std}, \textit{norm\_targ\_strg}, and \textit{norm\_res}, which normalizes on the residual of the K-nearest neighbours. 

\section{Experimental results}
The following is a limited baseline comparison of Target Strangeness with other nearest neighbour-based difficulty estimators implemented in the CREPES CP framework \citet{crepes}. Finally uncertainty is quantified for a Machine-Learning-Based Path Loss (MLPL) regression model using different difficulty estimators, and the results are compared.

\subsection{Baseline comparison}
The baseline was modified from the use case described in the \citet{crepes} journal paper. This problem involves creating prediction intervals for house sales. The baseline modified crepes\_nb notebook from version 0.5 of the CREPES \citet{bostrom2023crepes} package, to implement Target Strangeness, refer to crepes\_nb\_targ\_strg.py in the arXiv anc/code directory.

The results can be found in Tables: \ref{tab:cps_results_targ_strg}, \ref{tab:cps_results_std} and \ref{tab:cps_results_var} with the highlighted columns of interest corresponding to the CPS with the smallest prediction interval. CRPS is a loss function that measures the predictive efficiency of predictive distributions \citet{vovk2020computationally}. Note the results are standardized values.

The results contain CRPS values that are very similar, however the effective mean and median created by Target Strangeness is slightly smaller than the CPS normalize with conformal regressors using difficulty estimators of variance and standard deviation(std). This trend continues as the confidence is increased.

This limited regression baseline demonstrates that Target Strangeness is a viable difficulty estimator able to capture local uncertainties, and comparable to others in the CREPES framework.

\begin{table}[h]
\centering
\begin{tabular}{lcccccc}
\hline
CPS & \multicolumn{3}{c}{cps\_norm\_targ\_strg} & \multicolumn{3}{c}{cps\_mond\_norm\_targ\_strg} \\
confidence & 0.9 & 0.95 & 0.99 & 0.9 & 0.95 & 0.99 \\
\hline
error & 0.1024 & 0.0503 & 0.0105 & 0.1006 & 0.0478 & 0.0053 \\
eff\_mean & \cellcolor{yellow!25}0.0404 & 0.0502 & 0.0710 & 0.0420 & 0.0510 & 0.0733 \\
eff\_med & \cellcolor{yellow!25}0.0273 & 0.0345 & 0.0505 & 0.0292 & 0.0359 & 0.0504 \\
CRPS & \cellcolor{yellow!25}0.0069 & 0.0069 & 0.0069 & 0.0069 & 0.0069 & 0.0069 \\
time\_fit & 0.0002 & 0.0002 & 0.0002 & 0.0003 & 0.0003 & 0.0003 \\
time\_evaluate & 0.7106 & 0.6670 & 0.6690 & 0.4990 & 0.5072 & 0.4958 \\
\hline
\end{tabular}
\caption{Results for CPS with CR normalized with Target Strangeness}
\label{tab:cps_results_targ_strg}
\end{table}

\begin{table}[h]
\centering
\begin{tabular}{lcccccc}
\hline
CPS & \multicolumn{3}{c}{cps\_norm\_std} & \multicolumn{3}{c}{cps\_mond\_norm\_std} \\
confidence & 0.9 & 0.95 & 0.99 & 0.9 & 0.95 & 0.99 \\
\hline
error & 0.1008 & 0.0474 & 0.0090 & 0.0989 & 0.0489 & 0.0093 \\
eff\_mean & 0.0455 & 0.0609 & 0.0987 & \cellcolor{yellow!25}0.0456 & 0.0603 & 0.1037 \\
eff\_med & 0.0354 & 0.0477 & 0.0792 & \cellcolor{yellow!25}0.0339 & 0.0449 & 0.0775 \\
CRPS & 0.0071 & 0.0071 & 0.0071 & \cellcolor{yellow!25}0.0070 & 0.0070 & 0.0070 \\
time\_fit & 0.0002 & 0.0002 & 0.0002 & 0.0003 & 0.0003 & 0.0003 \\
time\_evaluate & 0.7428 & 0.7194 & 0.6612 & 0.5407 & 0.5361 & 0.5033 \\
\hline
\end{tabular}
\caption{Results for CPS with CR normalized with standard deviation}
\label{tab:cps_results_std}
\end{table}

\begin{table}[h]
\centering
\begin{tabular}{lcccccc}
\hline
CPS & \multicolumn{3}{c}{cps\_norm\_var} & \multicolumn{3}{c}{cps\_mond\_norm\_var} \\
confidence & 0.9 & 0.95 & 0.99 & 0.9 & 0.95 & 0.99 \\
\hline
error & 0.0984 & 0.0492 & 0.0103 & 0.1059 & 0.0511 & 0.0064 \\
eff\_mean & 0.0464 & 0.0590 & 0.0886 & \cellcolor{yellow!25}0.0449 & 0.0591 & 0.0983 \\
eff\_med & 0.0263 & 0.0338 & 0.0528 & \cellcolor{yellow!25}0.0276 & 0.0348 & 0.0590 \\
CRPS & 0.0070 & 0.0070 & 0.0070 & \cellcolor{yellow!25}0.0068 & 0.0068 & 0.0068 \\
time\_fit & 0.0002 & 0.0002 & 0.0002 & 0.0003 & 0.0003 & 0.0003 \\
time\_evaluate & 0.6673 & 0.6691 & 0.6675 & 0.4944 & 0.4961 & 0.5014 \\
\hline
\end{tabular}
\caption{Results for CPS with CR normalized with variance}
\label{tab:cps_results_var}
\end{table}

\newpage

\subsection{Machine-Learning-Based Path Loss (MLPL) regression model}
The MLPL regression task involved generating PIs for a machine learning-based wireless path loss model, using external features. Path loss is the decrease in wireless signal strength as it propagates, influenced by factors such as frequency, distance and obstacles. In designing communications systems, one frequently prioritizes having high levels of confidence in a particular metric, so PIs may be more important than simple mean or mode estimation. This experiment was used to determine whether CP PIs can be generated for a path loss model and whether the Target Strangeness difficulty estimator is comparable to other estimators.
\subsubsection{Data preparation}
\label{sec:MLPL_data_prep}
This machine-learning-based path loss model by \cite{chateauvert2024machine} was trained on over 5.7 million samples of open U.K. path loss drive test data \cite{OfcomDrivetest}, and consists of three external input features for single target regression. Two of the features (frequency and distance) are available in the open data sets, and an additional tool \cite{SAFE} was used to extract the total obstacle depth feature from open digital terrain models (DTM) \cite{UK_DTM} and open digital surface models (DSM) \cite{UK_DSM} elevation data. The CP calibration set was randomly sampled and removed from the training set, before model training. The trained model was used to generate predictions for the train set, CP calibration set and test set. 

These three CP datasets were then randomly sampled, retaining approximately 1\% of the original data to reduce the CP computation time. The training set was all of the U.K. with the London drive test held out as the blind test set. The initial dataset details are in Table \ref{tab:mlpl_dataset_counts}.

\begin{table}[htbp]
\caption{CRC MLPL dataset counts and descriptions}
\centering
\begin{tabular}{lll}
\toprule
\textbf{Dataset} & \textbf{Count} & \textbf{Description} \\
\midrule
train & 4171745 & UK-not London \\
calibrate & 1042937 & UK-not London \\
test & 582260 & London \\
\midrule
external features & 3 & Frequency, Link Distance, Total Obstacle Depth \\
\bottomrule
\end{tabular}
\label{tab:mlpl_dataset_counts}
\end{table}

\subsubsection{Machine-Learning-Based Path Loss (MLPL) model architecture}
Various curve-fitting and ML-based models were considered by \cite{chateauvert2024machine}. We focus on a \cite{XGBoost} boosted tree model consisting of 200 trees, each with a maximum depth of 3, trained with a learning rate of 0.04. The model was trained in minutes. 
\subsubsection{MLPL prediction interval results}
The following results were obtained by performing a grid search for two hyperparameters: the number of K-nearest neighbours and the number of Mondrian bins (ranging from 10 to 100 in increments of 10). This search was conducted for the five outlined difficulty estimators and five different seeds. The hyperparameters yielding the narrowest PI width were chosen for each estimator. The effective coverage is the percentage of predictions that fall within the PIs. To ensure comparability, results were analyzed at approximately a 90\% effective coverage. 
The experiments requested PIs with a 90\% and 95\% confidence and took the means around the effective coverage of interest. 

\begin{figure}[htbp]
    \centering
    \includegraphics[width=0.7\textwidth]{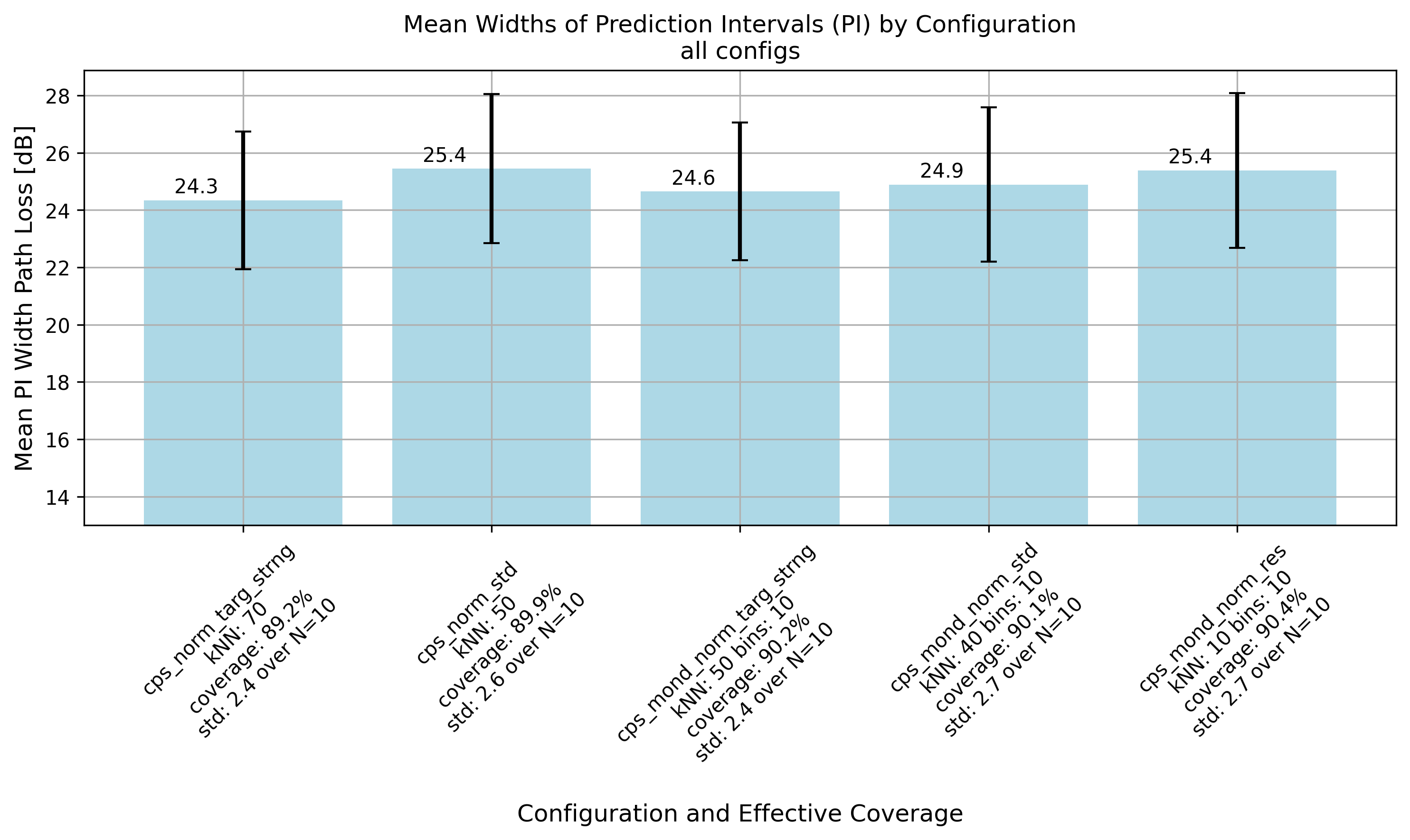}
    \caption{MLPL experiments}
    \label{fig:MLPL_experiment}%
\end{figure}

\begin{figure}[htbp]
    \centering
    \includegraphics[width=0.7\textwidth]{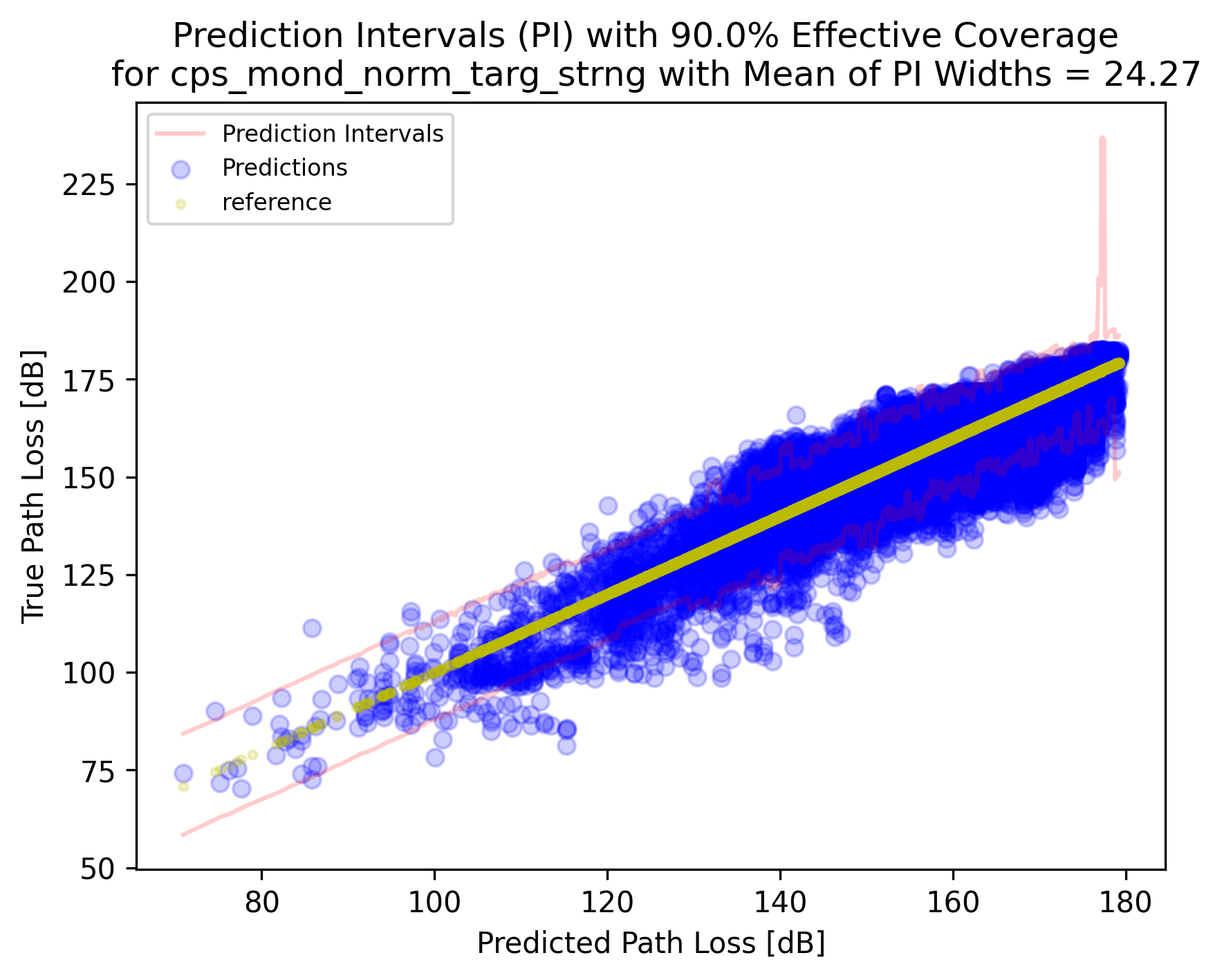}
    \caption{MLPL PIs result example}
    \label{fig:MLPL_experiment_example}%
\end{figure}

The following experimental configuration and effective coverage plot, Figure \ref{fig:MLPL_experiment}, contain the number of aggregated samples used to calculate the means and respective standard deviation, also represented as error bars. These results are summarized in Table \ref{tab:mlpl_results}. This table contains the data features and CP method description, mean PI width and finally the optimal K-nearest neighbours (kNN), and Mondrian bins where applicable. An example of one prediction interval configuration is shown in \ref{fig:MLPL_experiment_example}.

The MLPL results in Table \ref{tab:mlpl_results} indicate that CP can be used to create PIs for path loss. All the mean PI widths appear to be similar; however, the decibel unit is log-scale (as a reference, an additional 3 dB is twice the power). Therefore, a small change in dB is meaningful. In the context of path loss modeling, the Target Strangeness method is observed to be a viable difficulty estimator as it is marginally better than all other methods studied. 

The prediction uncertainty is affected by measurement uncertainty related to input features, epistemic uncertainty caused by the possible inadequacy of features fed into the model and aleatoric uncertainty tied to output predictions. These various uncertainty sources contribute to the 24.3 dB total uncertainty interval result for MLPL. This demonstrates that Target Strangeness can be a viable difficulty estimator option for path loss regression problems.

\begin{table}[htbp]
  \caption{MLPL prediction interval results}
  \label{tab:mlpl_results}
  \centering
  \begin{tabular}{llccc}
    \toprule
    \textbf{Configuration} & \textbf{Mean PI} & \textbf{Effective}\\
    \textbf{(data and CP method)} & \textbf{Width [dB]} & \textbf{Coverage [\%]} & \textbf{kNN} & \textbf{Bins}\\
    \midrule
    \multicolumn{5}{l}{\textbf{External Tabular Features}} \\
    norm\_std & 25.4 & 89.9 & 50 & \\
    \cellcolor{green!30}norm\_targ\_strg & \cellcolor{green!30}24.3 & \cellcolor{green!30}89.2 & \cellcolor{green!30}70 & \\
    \midrule
    \multicolumn{5}{l}{\textit{Mondrians}} \\
    norm\_std & 24.9 & 90.1 & 40 & 10 \\
    norm\_res & 25.4 & 90.4 & 10 & 10 \\
    norm\_targ\_strg & 24.6 & 90.2 & 50 & 10 \\
    \bottomrule
  \end{tabular}
\end{table}

\newpage
\section{Limitations}

The effectiveness of a normalized conformal nonconformity estimator is inherently shaped by several factors, including the data's intrinsic properties, the chosen normalization scheme, and the underlying assumptions inherent in the conformal prediction framework. Consequently, empirical validation, alongside comparison with alternative approaches like the Target Strangeness difficulty estimator, is crucial to assess its practical utility and robustness.

\section{Conclusions}

This paper introduced the Target Strangeness difficulty estimator which uses a kernel density estimate to quantify the unlikeliness of the prediction given the distribution of the nearest neighbours. The stranger a prediction is relative to the target distribution, the higher the uncertainty, which results in larger prediction intervals (PIs). The Target Strangeness difficulty estimator was integrated within the CREPES framework and compared against other difficulty estimators in a baseline, and in the context of wireless path loss modeling. Empirical results revealed that Target Strangeness outperformed other difficulty estimators, underscoring its potential to enhance conformal prediction intervals. However, further investigation remains necessary to fully explore its capabilities.

\section{Acknowledgements}
I extend my deepest gratitude to my father, Amitava \citet{BoseAmitava}, for his support, and to my grandfather, Asok \citet{BoseAsok} \citet{BoseAsokMemorial2006}, whose scientific legacy inspires me. I also honor my great-grandfather, Akshay Bose, who graduated from the now Visva-Bharati under Rabindranath Tagore in 1908, shaping our family's academic path.

\bibliographystyle{plainnat}
\bibliography{Styles/references}

\mbox{} 

\end{document}